\documentclass[letterpaper, 10 pt, conference]{ieeeconf}  

\IEEEoverridecommandlockouts                              

\title{\LARGE \bf
End-to-End RGB-IR Joint Image Compression With Channel-wise Cross-modality Entropy Model
}

\author{
\normalsize
Haofeng Wang$^{1,2,5}$, Fangtao Zhou$^{3}$, Qi Zhang$^{2}$, Zeyuan Chen$^{2,4}$,
Enci Zhang$^{1}$, Zhao Wang$^{5,2}$, \\ \\Xiaofeng Huang$^{3*}$, Siwei Ma$^{2*}$%
\thanks{* Corresponding authors.}%
\thanks{$^{1}$School of Electronic and Computer Engineering, Peking University, Shenzhen, China}%
\thanks{$^{2}$National Engineering Research Center of Visual Technology, School of Computer Science, Peking University, Beijing, China}%
\thanks{$^{3}$School of Communication Engineering, Hangzhou Dianzi University, Zhejiang, China}%
\thanks{$^{4}$Pengcheng Laboratory, Shenzhen, China}%
\thanks{$^{5}$Advanced Institute of Information Technology, Peking University, Zhejiang, China}%
}

\usepackage{multirow}
\usepackage{booktabs}
\usepackage{amsmath}
\usepackage{graphicx}

\begin{document}

\maketitle
\thispagestyle{empty}
\pagestyle{empty}

\begin{abstract}

RGB-IR(RGB-Infrared) image pairs are frequently applied simultaneously in various applications like intelligent surveillance. However, as the number of modalities increases, the required data storage and transmission costs also double. Therefore, efficient RGB-IR data compression is essential. This work proposes a joint compression framework for RGB-IR image pair. 
Specifically, to fully utilize cross-modality prior information for accurate context probability modeling within and between modalities, we propose a Channel-wise Cross-modality Entropy Model (CCEM). Among CCEM, a Low-frequency Context Extraction Block (LCEB) and a Low-frequency Context Fusion Block (LCFB) are designed for extracting and aggregating the global low-frequency information from both modalities, which assist the model in predicting entropy parameters more accurately. Experimental results demonstrate that our approach outperforms existing RGB-IR image pair and single-modality compression methods on LLVIP and KAIST datasets. For instance, the proposed framework achieves a 23.1\% bit rate saving on LLVIP dataset compared to the state-of-the-art RGB-IR image codec presented at CVPR 2022.

\end{abstract}

\section{INTRODUCTION}

Recently, RGB-IR images pairs captured within the same scene have been jointly applied to various practical scenarios\cite{b1,b2,b3}. This is largely due to the fact that the advantages of RGB and IR modalities are complementary. RGB images, known for their high resolution and ability to capture fine details such as textures, are limited by the reliance on ambient lighting\cite{b4}. However, this limitation can be mitigated by incorporating IR images because of the low sensitivity to illumination changes. Nevertheless, the use of RGB-IR image pairs significantly increases the amount of data that needs to be transmitted and stored. Consequently, developing an efficient joint compression method for RGB-IR image pairs has become a crucial and challenging task.

Over the past decades, deep learning-based image compression methods\cite{b5,b6,b7,b8,b9,b10} have been extensively developed, pushing the boundaries of rate-distortion performance. It is intuitive to compress RGB and IR modalities independently using these neural codecs. However, the redundancy between RGB and IR modalities is not fully exploited during the compression, thereby limiting the overall rate-distortion performance.

In recent years, several multi-modality data compression methods\cite{b11,b12} have been proposed. However, most of these methods are specifically designed for visible images paired with depth or hyperspectral images, which are not suitable for compressing RGB-IR image pairs due to the different distributions between modalities. For example, unlike depth images that use spatial geometry, IR images capture thermal properties and are less sensitive to lighting. For RGB-IR image pairs, a learning-based multimodal image compression framework\cite{b13} leverages one modality as an anchor to assist in the encoding and decoding process of the other modality. While this approach enhances the compression performance of one modality, it does not leverage the cross-modality correlation in the context-based entropy model, thereby limiting the rate-distortion performance of both modalities, which is often necessary in practical applications where RGB-IR image pairs are used together\cite{b14,b15}. Besides, the compression of two modalities cannot be performed simultaneously, as one modality has to be decoded at first to serve as an anchor for compressing the other, which lowers the computation efficiency. Therefore, designing a framework capable of jointly compressing RGB-IR image pairs by fully exploiting cross-modality correlations as prior information to enhance performance remains a challenge.

In this paper, our main contribution is to propose a dual-branch learning-based RGB-IR joint image compression framework to simultaneously compress RGB-IR image pairs, leveraging the correlation between modalities to save bit rate. We design a Channel-wise Cross-modality Entropy Model (CCEM) to fully utilize cross-modality prior information for accurate context probability modeling within and between modalities. Within CCEM, we propose Low-frequency Context Extraction Block(LCEB) and Low-frequency Context Fusion Block(LCFB) to extract and aggregate low-frequency prior information to further reveal the dependency between the modalities. Besides, unlike previous learning-based method for RGB-IR image pair compression, our approach does not require decoding one modality's image to be an anchor for compressing another. According to the experimental results, our proposed framework attains state-of-the-art performance compared to existing RGB-IR image pair and single-modality compression methods on LLVIP \cite{b16} and KAIST datasets \cite{b17}.

\begin{figure*}[htbp]
\centerline{\includegraphics[width=0.90\textwidth, height=0.36\textheight]{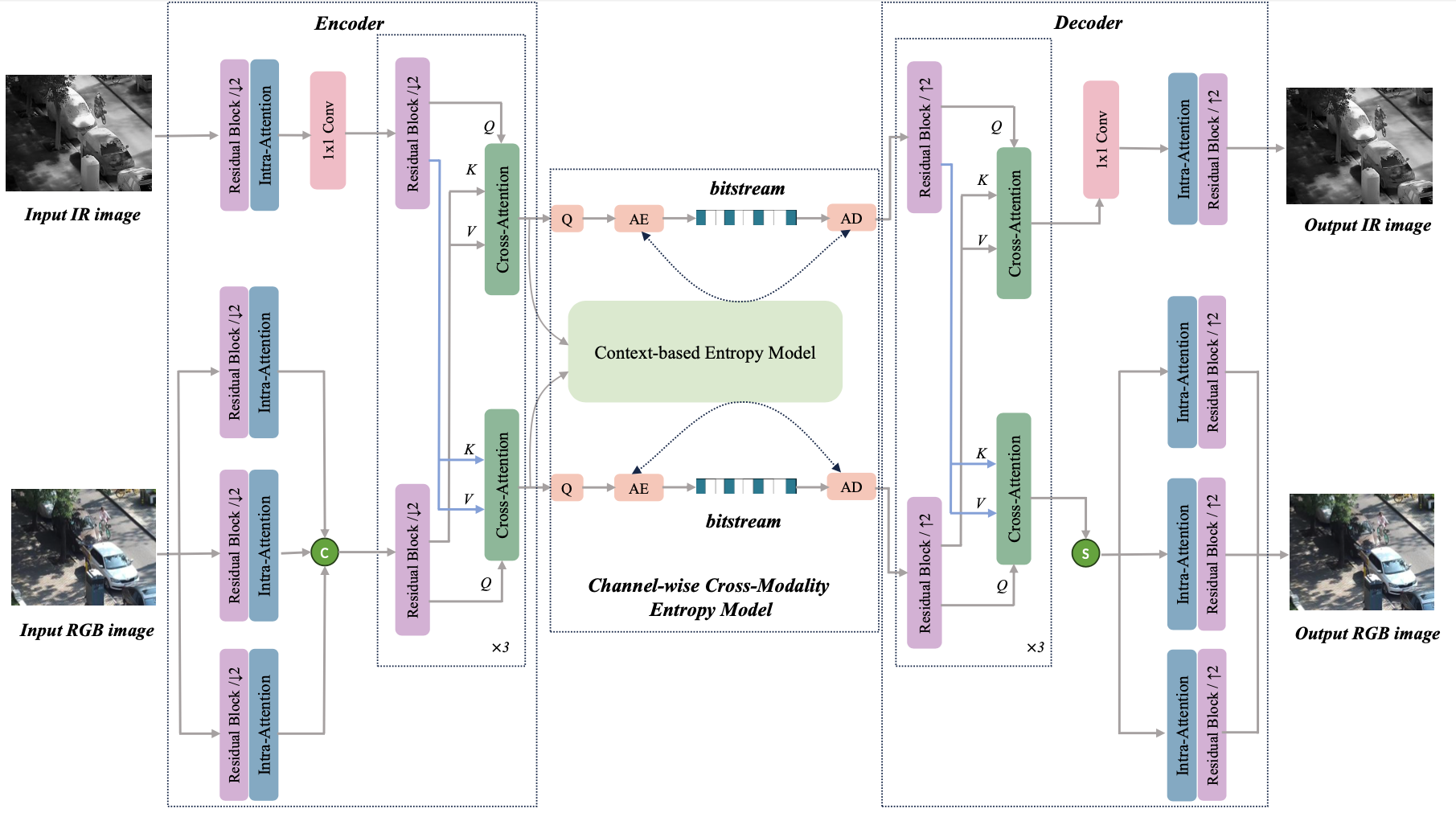}}
\caption{The overall framework of the proposed method. The network consists of an encoder, a Channel-wise Cross-Modality Entropy Model and a decoder. AE, AD denote arithmetic encoding and decoding, respectively. Q denotes the quantizer, C and S denote concat and split operation, "$\uparrow 2$" and "$\downarrow 2$" denote upsampling and downsampling by a factor of two, respectively.}
\label{fig1}
\vspace{-0.5cm} 
\end{figure*}

\section{PROPOSED METHOD}
\subsection{Overall Architecture}
The overall architecture of our RGB-IR joint compression framework is illustrated in Fig.~\ref{fig1}. We use a transformer-based encoder-decoder architecture. Before compression, the RGB image is converted to the YUV420 format, and the Y, U, V, and IR channels are used as inputs of the model. First, the input channels are individually fed into the Encoder for feature extraction. 
We use a residual network\cite{b6} combined with a self-attention-based module \cite{b18} to obtain feature maps $y^y$, $y^u$, $y^v$, and $y^{ir}$ for each input channel. The feature maps from the Y, U, and V channels are then concatenated to form a unified YUV feature $y^{yuv}$. We use cross-attention to embed cross-modality information within the latent representations $y^{yuv}$ and $y^{ir}$. Subsequently, $y^{yuv}$ and $y^{ir}$ are quantized to $\hat{y}^{yuv}$ and $\hat{y}^{ir}$, and fed into the proposed Channel-wise Context-based Cross-modality Entropy Model for accurate symbol probability prediction. Finally, $\hat{y}^{yuv}$ and $\hat{y}^{ir}$ are input into the decoder for upsampling and image reconstruction. We denote the encoder, quantizer, decoder as $g_a(\cdot)$, $Q(\cdot)$, and $s_a(\cdot)$, respectively. The overall process can be formulated as:
\begin{equation}
y^i=g_a(x^i;\theta),\hat{y^i}=Q(y^i),\hat{x^i}=g_s(\hat{y^i};\phi)
\end{equation}
where $x^i$ and $\hat{x^i}$ represents one of the input and output channels and $\theta$, $\phi$ are learnable parameters.

\subsection{Channel-Wise Cross-modality Entropy Model}
The entropy model plays a key role in boosting compression performance by estimating the distribution of the latent representation. Minnen et al.\cite{b19} introduced an entropy model based on spatial autoregressive prediction, surpassing the compression performance of H.265. To accelerate decoding, another work\cite{b20} have proposed to split the latent representation into multiple slices and leveraging inter-channel correlations to autoregressively predict the entropy model parameters for each slice. Based on this, MLIC++\cite{b10} incorporates multiple perspectives of context information as multi-references to predict entropy model parameters more accurately. For RGB-IR image pairs, leveraging cross-modality information, which has not been fully utilized, as prior context to enhance the accuracy of entropy model 
\begin{figure}[htbp]
\centerline{\includegraphics[width=0.3\textwidth, height=0.34\textheight]{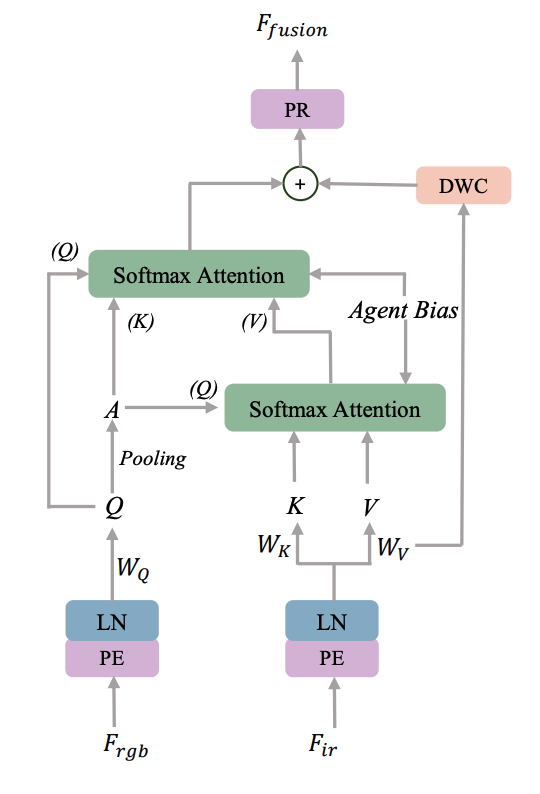}}
\caption{The architecture of Low-frequency Context Fusion Block(LCFB). PE, PR, LN represent Patch Embedding, Patch Recovery, LayerNorm, respectively.}
\label{fig3}
\vspace{-0.5cm} 
\end{figure}
\begin{figure*}[htbp]
\centerline{\includegraphics[width=1.0\textwidth, height=0.23\textheight]{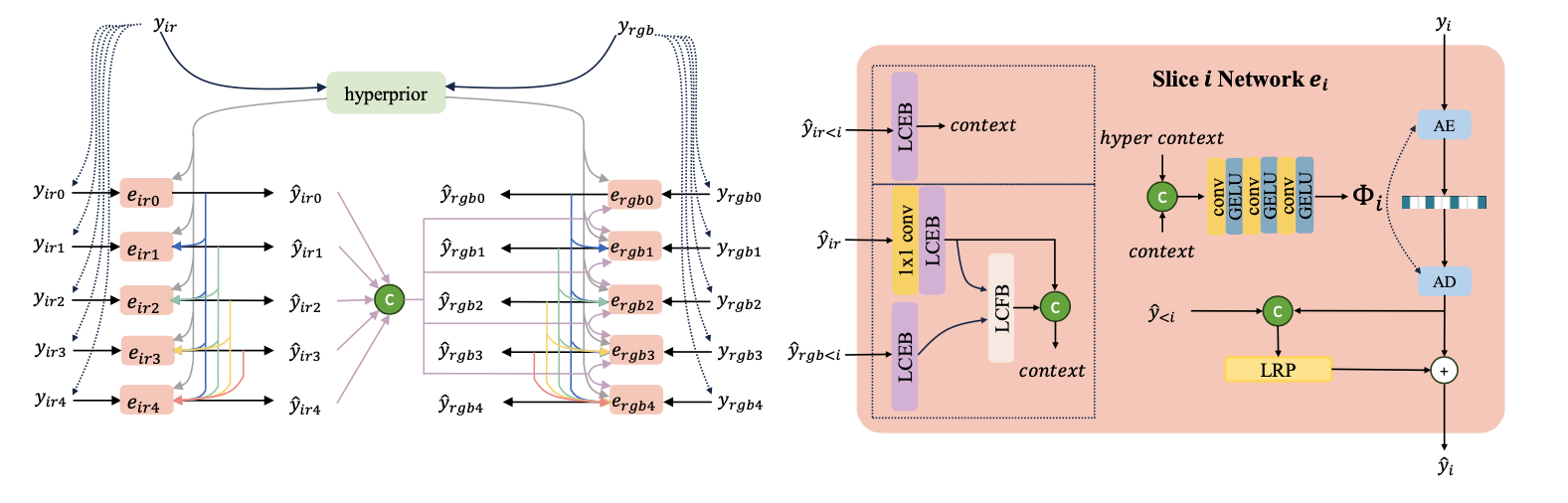}}
\vspace{-0.3cm}  
\caption{The architecture of the proposed Channel-wise Cross-Modality Entropy Model. The latent representations are split into slices and sent to hyperprior model. The encoded slices are fed into Low-frequency Context Extraction Block (LCEB) and Low-frequency Context Fusion Block (LCFB) to extract global low-frequency prior, then in slice entropy model $e_i$, hyperprior context and global low-frequency context are used to predict entropy parameters. LRP represents latent residual prediction module. C denotes concatenate operation.}
\label{fig2}
\vspace{-0.5cm} 
\end{figure*}
parameters prediction is a natural and worthwhile problem to explore.

The global low-frequency information of RGB images and IR images from the same scene is highly similar\cite{b21}. Therefore, it is reasonable to infer that, in the compression of RGB-IR image pairs, extracting and aggregating the global low-frequency information from both modalities as a conditional prior will enable the context-based entropy model to predict the parameters of the entropy model more accurately, thereby effectively reducing the bit rate. To verify this, We design the Low-frequency Context Extraction Block (LCEB) and Low-frequency Context Fusion Block (LCFB). The role of LCEB is to extract global low-frequency information within the modality, and since low-frequency information is typically distributed over large regions, it requires cross-regional global information exchange. Therefore, we adopt the Lite Transformer architecture\cite{b21}, as it can model long-range dependencies globally, making it particularly suitable for capturing low-frequency information. As shown in Fig.~\ref{fig3}, Instead of using a concatenation operation, we designed the LCFB based on agent-attention \cite{b22} to better aggregate global low-frequency information from two modalities. As agent-attention allows dynamic and selective weighting of information from both modalities, it enables more effective and context-aware fusion of low-frequency features, thereby enhancing the prediction of cross-modal entropy model parameters. The pipeline for processing the latent representations of the two modalities through the LCFB is as follows: 
\begin{equation}
\small
\begin{aligned}
    \{Q, K, V\} &= \{F_{\text{rgb}} \mathbf{W}^Q,\ F_{\text{ir}} \mathbf{W}^K,\ F_{\text{ir}} \mathbf{W}^V\}, \\
    A &= \mathrm{Pooling}(Q), \\
    V' &= \mathrm{softmax}\left( \frac{A K^\top}{\sqrt{d_k}} + B_1 \right) V, \\
    F &= \mathrm{softmax}\left( \frac{Q A^\top}{\sqrt{d_a}} + B_2 \right) V', \\
    F_{\text{fusion}} &= F + \mathrm{DWC}(V).
\end{aligned}
\label{eq:fusion}
\end{equation}

where $F_{rgb}$ and $F_{ir}$ represent feature of input slices. $\mathbf{W}^{Q}$, $\mathbf{W}^{K}$, $\mathbf{W}^{V}$ are linear projection matrices to map the input features into query $Q$, key $K$, and value $V$ spaces, respectively. $d$, $B$ represent the dimension and relative position bias , respectively. DWC is a depth-wise convolution module\cite{b23}. We see the output $F_{fusion}$ as the aggregated global low-frequency information from two modalities and use this context in entropy model to get more accurate entropy parameters.

Combining LCEB and LCFB, we design a Channel-wise Cross-modality Entropy Model (CCEM) for more accurate probability estimation. The architecture of CCEM is shown on the left side of Fig.~\ref{fig2}. The latent representation generated from the encoder is fed into a hyperprior model to obtain spatial prior information. Additionally, the latent representation is divided into slices $\{\hat{\boldsymbol{y_m}}^0,\hat{\boldsymbol{y_m}}^1,\cdots\hat{\boldsymbol{y_m}}^N\}$, where $m$ represents one of input modalities. For the IR latent representation, the first slice uses only the hyperprior as context to predict entropy model parameters. For the $i^{th}$ slice, we use the previous slices to exact context and predict entropy parameters. In particular, the slices from $1$ to $i-1$ are concatenated and processed through a LCEB to extract global low-frequency information. The global low-frequency context and hyperprior context are then used to predict entropy parameters. For the RGB latent representation, in addition to the above, the $j^{th}$ slice is processed by concatenating the preceding $j-1$ slices with the global low-frequency information from the previously obtained IR latent representation and input into a LCFB to derive cross-modality information. This additional cross-modality information is used to further improve the accuracy of the entropy model parameters prediction. Specifically, we denote $\hat{y}_{ir}$ and $\hat{y}_{r}$ as the latent representation of two modalities. $\hat{z}$ represents the side information extracted from hyperprior. The probability distribution of the latent variables $p_{\hat{y}_{ir}}$ and $p_{\hat{y}_{r}}$ can be formulated as:
\begin{equation}
\fontsize{10pt}{6pt}\selectfont 
\begin{aligned}
p_{\hat{y}_{ir} | \hat{z}_{ir}} (\hat{y}_{ir} | \hat{z}_{ir}) &= \prod_{i=1}^{N} p_{\hat{y}_{ir}^i | \hat{y}_{ir}^{<i}, \hat{z}_{ir}} (\hat{y}_{ir}^i | \hat{y}_{ir}^{<i}, \hat{z}_{ir}), \\
p_{\hat{y}_r | \hat{y}_{ir}, \hat{z}_r} (\hat{y}_r | \hat{y}_{ir}, \hat{z}_r) &= \prod_{i=1}^{N} p_{\hat{y}_r^i | \hat{y}_r^{<i}, \hat{y}_{ir}, \hat{z}_r} (\hat{y}_r^i | \hat{y}_r^{<i}, \hat{y}_{ir}, \hat{z}_r).
\end{aligned}
\end{equation}

\begin{figure*}[htbp]
    \centering
    \begin{minipage}{0.48\textwidth}  
        \centering
        \includegraphics[width=1.1\textwidth]{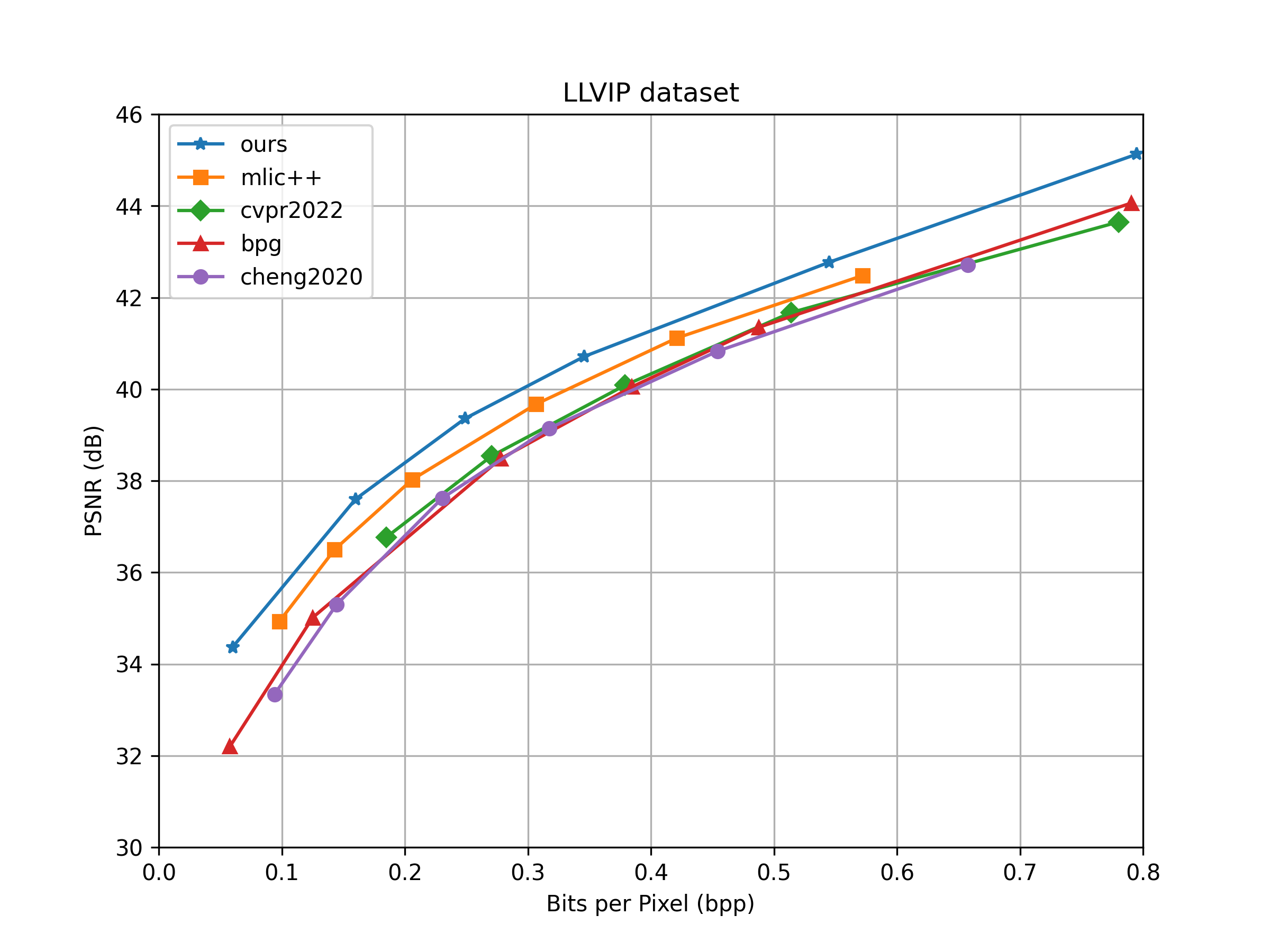}  
    \end{minipage}\hspace{0.1cm}  
    \begin{minipage}{0.48\textwidth}  
        \centering
        \includegraphics[width=1.1\textwidth]{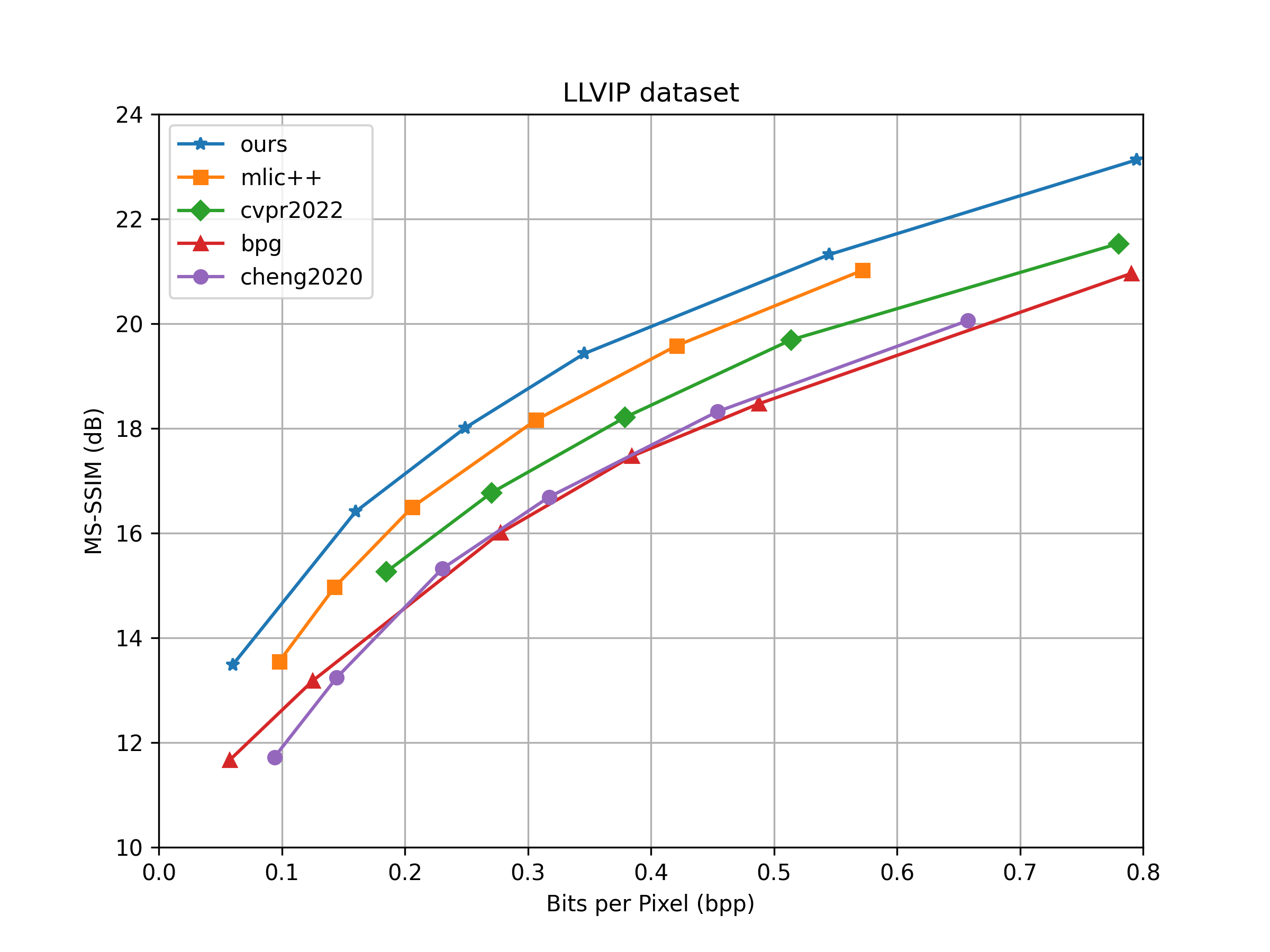}  
    \end{minipage}\\[0.3cm]  
    \begin{minipage}{0.48\textwidth}  
        \centering
        \includegraphics[width=1.1\textwidth]{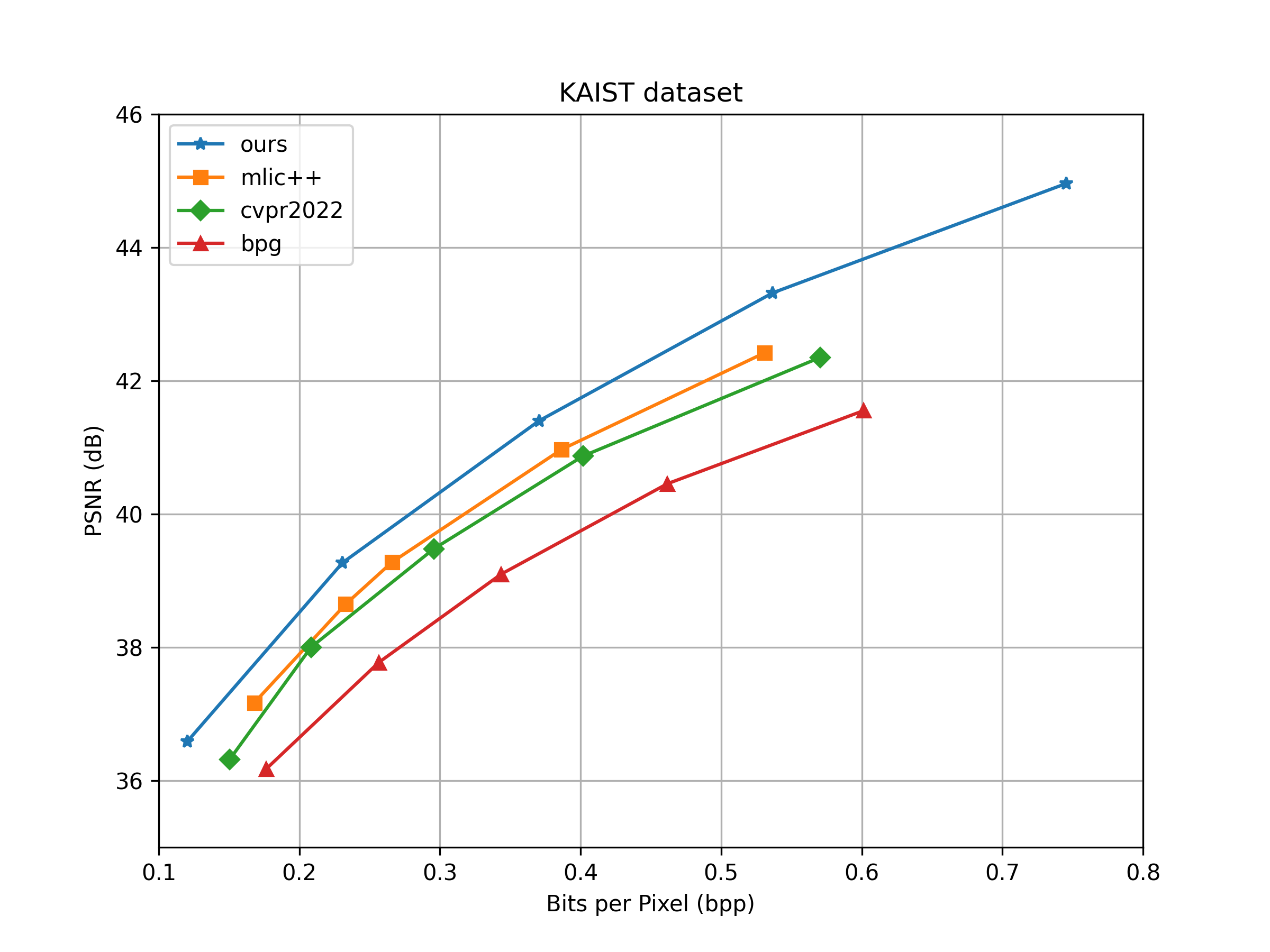}  
    \end{minipage}\hspace{0.1cm}  
    \begin{minipage}{0.48\textwidth}  
        \centering
        \includegraphics[width=1.1\textwidth]{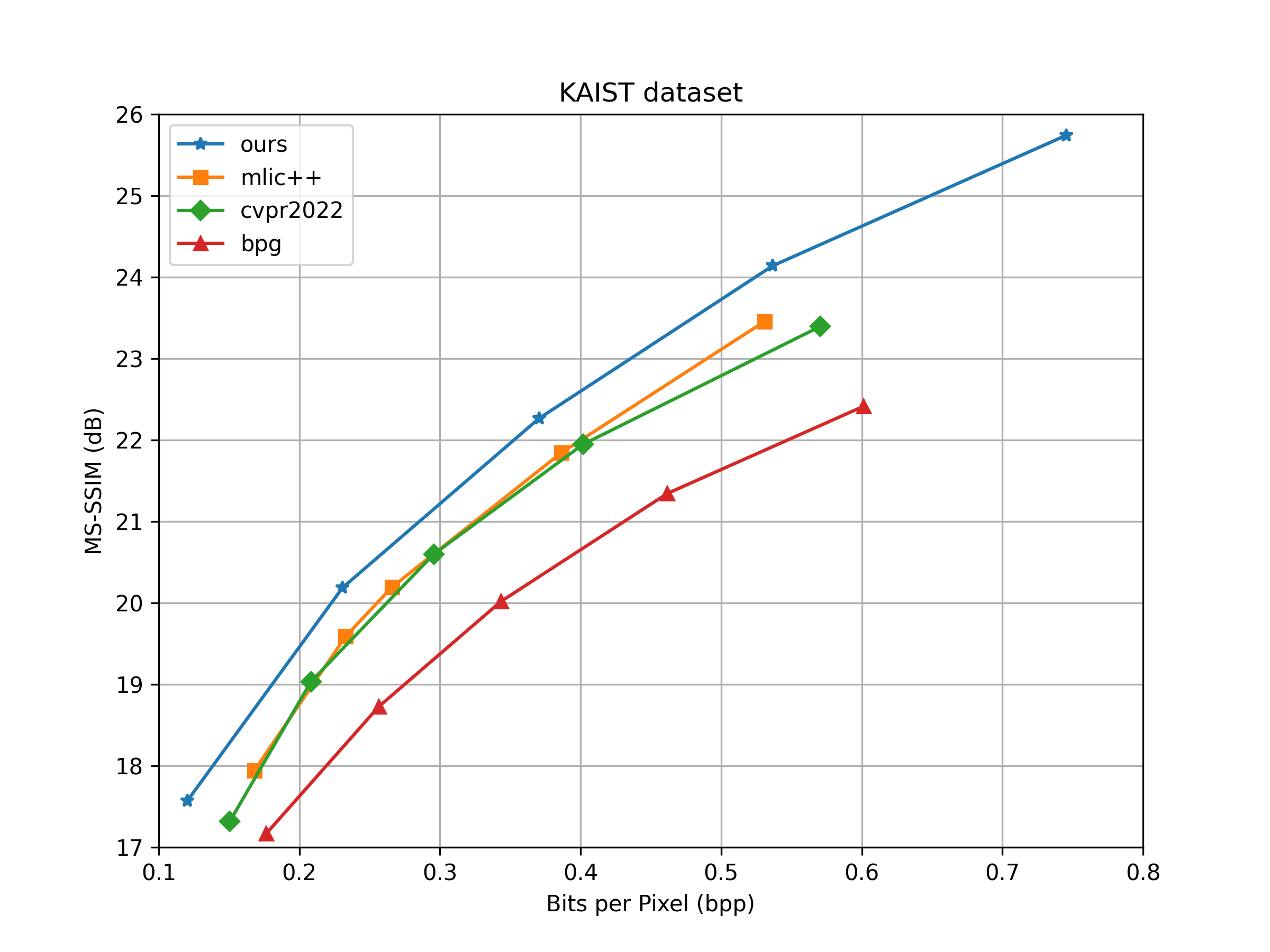}  
    \end{minipage}

    \caption{Experimental results from different image compression approaches on the LLVIP and KAIST datasets.}
    \label{fig4}
\end{figure*}

\subsection{Loss Function}
The loss function $L$ of our framework is described as: 
\begin{equation}L=R_{ir}+R_{rgb}+\lambda(D_{ir}+D_{rgb}).\end{equation}
where $R_{ir}$ and $R_{rgb}$ are the bit rate cost of two modalities, they can be calculated by the probability distribution of latent representations. $D_{rgb}$ and $D_{ir}$ are calculated as the pixel-wise mean square error (MSE) between compressed and original image.
\section{Experiments}
\subsection{Experiment Details}
\textbf{Baseline and Metric} \ We introduce state-of-the-art RGB-IR image compression method (Hereafter, referred to as CVPR2022)\cite{b13}, for comparison with our model. Additionally, we compare our model with the best-performing single-modality codec on the Kodak dataset, MLIC++\cite{b10}, and the classic end-to-end codec, Cheng2020\cite{b6}, traditional single-modality image compression method BPG\cite{b24}. To ensure a fair comparison, we fine-tuned all end-to-end compression methods on the LLVIP and KAIST datasets. For single-modality codecs, which are primarily trained on RGB-IR images, we duplicated the single-channel IR images into three channels to preserve the original model structure during training with IR images. This approach follows the methods used in previous learning-based multimodality compression studies\cite{b12}\cite{b13}, where existing codecs were directly fine-tuned using IR images. Additionally, for the BPG codec, when encoding IR images, we set the pixel\_format option to 0 (Grayscale). We use PSNR and MS-SSIM to assess the quality of compressed images and Bjontegaard delta rate (BD-Rate)\cite{b25} to evaluate the rate-distortion performance. Considering our ultimate goal is the joint compression of RGB-IR image pairs, we compare the average PSNR of both modalities and the corresponding BD-Rate with the baseline models. Note that all evaluation metrics are computed in the YUV420 domain.

\begin{figure*}[htbp]
\centerline{\includegraphics[width=1.0\textwidth, height=0.209\textheight]{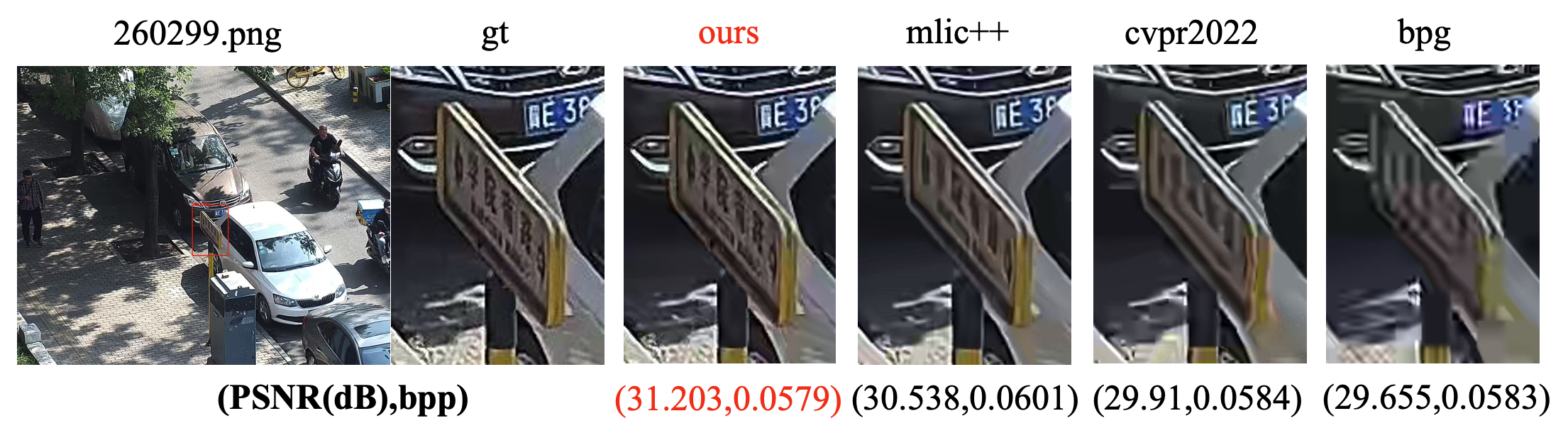}}
\caption{Visualization of reconstructions of 260299.png in LLVIP using different methods.}
\label{fig5}
\vspace{-0.5cm}
\end{figure*}

\textbf{Training details} \ Considering that joint training of both modalities from the beginning would require the model to simultaneously process multiple channels from two modalities, it's difficult to learn the features of each modality and their cross-modality correlations. During model training, we propose a two-stage training method. In the first stage, we focus on training for compressing the RGB data. Specifically, after converting the RGB modality to YUV, we input the Y channel data into the proposed model for training. This approach ensures that the model can effectively extract features from the RGB modality in the early stages. After completing the first stage, we proceed to jointly optimize both the RGB and IR modalities. Experimental results show that adopting this training method improves the model's performance by approximately 4\%. Additionally, we set different hyperparameters $\lambda$, to control the bit rate, following the settings in CompressAI \cite{b26}. During training, we use the Adam optimizer, and the learning rate gradually decreases from 1e-4 to 1e-5 throughout each stage. 

We conduct training and testing on LLVIP \cite{b16} and KAIST Pedestrian \cite{b17}, two widely used RGB-IR datasets. For LLVIP, Training is performed on the dataset's 12,000+ training images for 150 epochs in each stage, and testing is carried out on its 3,400+ pairs of test images. On the KAIST Pedestrian dataset, we use set00 to set05 for training and the rest for testing. Both datasets we tested were collected under different \textbf{lighting conditions} (from day through night). Additionally, to our knowledge, the LLVIP dataset is currently the \textbf{highest resolution} (1280x1024) RGB-IR image pair dataset. The KAIST dataset contains a significant number of \textbf{high-dynamic scenes} because it includes many fast-moving objects in real-time traffic scenes.

During the two-stage training process, we followed the training schemes of [26], setting the values of \(\lambda\) to \([0.0018, 0.0035, 0.0067, 0.0130, 0.0250, 0.0483]\). We followed \cite{b13}, treating the two modalities as equally important. Therefore, we did not introduce an additional weight in the loss function to balance the distortion between the two modalities. In the training process, images are randomly cropped to a size of \(256 \times 256\) and the batch-size is set to 4. The number of channels in the latent representation of each modality output by the Encoder is 320. Additionally, in the CCEM, the latent representation is divided into \(N = 5\) blocks. The structure of the hyperprior is consistent with that in \cite{b12}.

The training goal of the initial stage training is to maintain the consistency of the overall model architecture and enhance the model’s learning capability for the RGB modality during the early stages of training, we temporarily exclude the IR modality in the first stage. Thus, the loss function in the initial stage is similar to that of the second stage, except that the terms \(R_{ir}\) and \(D_{ir}\) from the second stage are removed, with all other hyperparameters remaining the same. 

\subsection{Experiment Results}
\textbf{Quantitative Results} \ 
We make a comparison of compression performance among various single-modality codecs and a state-of-the-art RGB-IR compression framework.  Compared to other single-modality compression frameworks, our proposed framework shows a significant improvement in BD-rate performance. Specifically, our method outperforms CVPR2022\cite{b13} and MLIC++\cite{b10} by 23.1\% and 14.6\%, respectively. We plot the corresponding RD curves in Fig.~\ref{fig4} to more intuitively illustrate the performance gap between different codecs. The results clearly demonstrate that our proposed method significantly outperforms the other methods in terms of compression performance.

\textbf{Qualitative Results} \
As depicted in Fig.~\ref{fig5}, our method exhibits superior subjective visual quality under the premise of using less bit rate. Specifically, after the local details are enlarged, our method can still retain the semantic information (such as the text on the roadside sign and the license plate number) of the original image.

\textbf{Modality-Specific Performance} \
As shown in Table~\ref{tab:comparison}, the proposed method achieves significant performance improvements for \textbf{both RGB and IR modalities}. This is due to the efficient utilization of low-frequency information from both modalities in the proposed CCEM, which makes the parameter estimation in the entropy model more accurate.

\begin{table}[h]
\vspace{+0.3cm}
\caption{Modality-Specific BD-Rate (\%) comparisons on LLVIP dataset and KAIST dataset against BPG.}
\label{tab:comparison}
\centering
\renewcommand{\arraystretch}{1.2} 
\begin{tabular}{l c c | c c}
\hline
\multirow{2}{*}{Methods} & \multicolumn{2}{c}{LLVIP} & \multicolumn{2}{c}{KAIST} \\
\cline{2-3} \cline{4-5}
 & RGB & IR & RGB & IR \\
\hline
Cheng2020 & 5.426 & -1.716 & 10.335 & -4.628 \\
CVPR2022 & -10.733 & -3.262 & -18.639 & -21.289 \\
MLIC++ & -9.463 & -18.034 & -16.761 & -25.622 \\
Ours & \textbf{-19.89} & \textbf{-35.051} & \textbf{-27.8236} & \textbf{-39.001} \\
\hline
\end{tabular}
\end{table}

\textbf{Running time and complexity} \
The model we proposed has 89.01M parameters, and compressing a RGB-IR image pair of size 1280x1024 requires 2.3GB of GPU memory. The FLOPs of the CCEM module reach 2.85 Mil/pixel. The model we proposed does not focus on real-time performance, but rather emphasizes Rate-Distortion (RD) performance. When tested on both datasets using a single NVIDIA 4090 machine, our model's average encoding time for an RGB-IR image pair is 881ms, and the average decoding time is 942ms. In comparison, prior sequential approaches, such as CVPR2022\cite{b13}, reach an encoding time of 697ms and a decoding time of 601ms. Anchor-based methods, like MLIC++\cite{b10}, reach an encoding time of 901ms and a decoding time of 978ms. While our method introduces slightly more running time costs, it achieves a significant improvement in RD performance. 

\textbf{Ablation Study}:
To demonstrate the effectiveness of the proposed LCEB and LCFB modules, we conducted experiments by removing each module individually and comparing the results. Table \ref{tab1} shows that both proposed modules contribute to BD-rate performance, and our proposed CCEM significantly enhances compression efficiency.

\begin{table}[htbp]
\caption{Ablation study of each component in Channel-wise Cross-modality Entropy Model}
\label{tab1}
\begin{center}
\begin{tabular}{|c||c|}
\hline
\textbf{Model} & \textbf{BD-Rate(\%)} \\
\hline
baseline & - \\
\hline
Channel-wise Cross-modality Entropy Model & -19.34 \\
\hline
baseline + LCEB & -6.92 \\
\hline
baseline + LCFB & -9.17 \\
\hline
\end{tabular}
\end{center}
\end{table}

\subsection{Other Exploration and Analysis}
In our CCEM, we extract contextual information from the decoded IR features through a series of transformations to assist in entropy parameters prediction for RGB features. In fact, we also conducted comparative experiments where RGB features were used to assist in entropy parameters prediction for IR features. Ultimately, on the LLVIP dataset, the former approach achieved approximately 8\% bit rate savings compared to the latter.

\begin{figure}[htbp]
\centerline{\includegraphics[width=0.5\textwidth, height=0.2357\textheight]{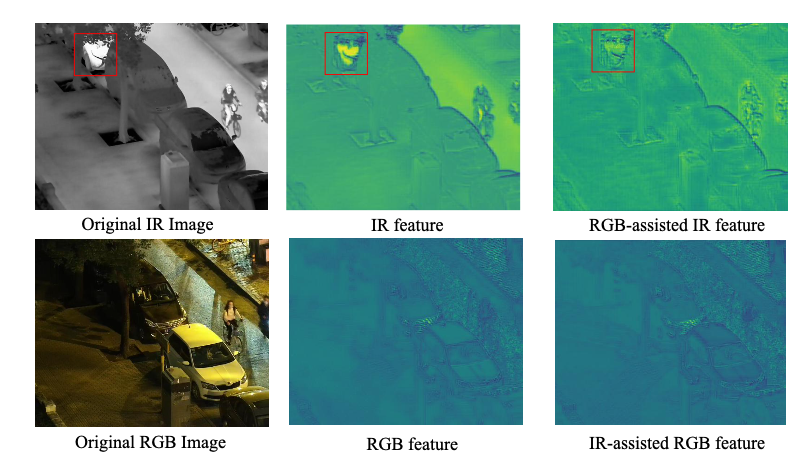}}
\caption{Visualization of the RGB and IR feature in CCEM }
\label{fig6}
\end{figure}

To investigate the underlying reasons, we visualized the IR and RGB features in the entropy models of the two approaches. As illustrated in the second line of Fig.~\ref{fig6}, the RGB features obtained with the assistance of decoded IR features exhibit more distinct structural information compared to the features without such assistance. Additionally, some regions experience texture enhancement, enabling more accurate entropy parameters prediction. This improvement is attributed to the low-frequency information extracted by our designed LCEB. Conversely, as illustrated in the first line of Fig.~\ref{fig6}, when using the decoded IR features to assist the RGB feature, the resulting feature contain richer texture information compared to the features without assistance. However, this approach also introduces noise in certain regions (as marked in the figure), which interferes with the accuracy of entropy parameters prediction.

\section{CONCLUSIONS}
In this paper, we propose a joint compression framework for RGB-IR image pair. Specifically, to remove cross-modality redundancy and save bit-rate, we introduce the Channel-wise Cross-modality Entropy Model (CCEM). Within CCEM, we design the Low-frequency Context Extraction Block (LCEB) and the Low-frequency Context Fusion Block (LCFB) based on the similarity of low-frequency information between RGB and IR images. These blocks effectively capture both intra-modality and cross-modality priors, thus assisting the entropy model in predicting symbol probability estimates more accurately. Comparative experiments and ablation studies confirm the effectiveness of the proposed method.





\addtolength{\textheight}{-12cm}   





\bibliographystyle{IEEEtran}
\bibliography{references}

\end{document}